\documentclass{article}

\usepackage{arxiv}

\usepackage[utf8]{inputenc} 
\usepackage[T1]{fontenc}    
\usepackage{hyperref}       
\usepackage{url}            
\usepackage{booktabs}       
\usepackage{amsfonts}       
\usepackage{nicefrac}       
\usepackage{microtype}      
\usepackage{lipsum}		
\usepackage{graphicx}
\usepackage{natbib}
\usepackage{doi}
\usepackage{multirow}

\title{Meta-Learning for One-Class Classification with Few Examples using Order-Equivariant Network}

\author{ Ademola Oladosu\thanks{Authors contributed equally to this work} \\
	New York University, Center for Data Science\\
	\And
	Tony Xu$^*$ \\
	New York University, Center for Data Science\\
	\And
	Philip Ekfeldt$^*$ \\
	New York University, Center for Data Science\\
	\And
	Brian A. Kelly$^*$ \\
	New York University, Center for Data Science\\
	\And 
	\href{https://orcid.org/0000-0002-6458-3423}{\includegraphics[scale=0.06]{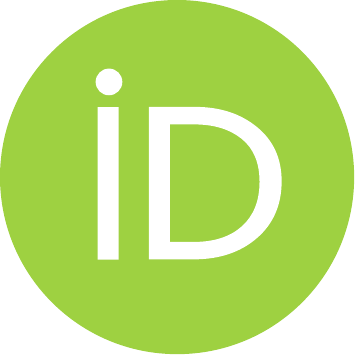}\hspace{1mm}Miles Cranmer}\\
	Princeton University \\
		\And
    Shirley Ho\\
	Center for Computational Astrophysics, Flatiron Institute\\
	\And
	\href{https://orcid.org/0000-0003-0872-7098}{\includegraphics[scale=0.06]{orcid.pdf}\hspace{1mm}Adrian M. Price-Whelan} \\
	Center for Computational Astrophysics, Flatiron Institute \\
	\And
	\href{https://orcid.org/0000-0002-3011-4784}{\includegraphics[scale=0.06]{orcid.pdf}\hspace{1mm}Gabriella Contardo}\\
	Center for Computational Astrophysics, Flatiron Institute \thanks{Corresponding author: \texttt{gcontardo@flatironinstitute.org} }
}

\renewcommand{\shorttitle}{Meta-Learning for One-Class Classification with Few Examples using Order-Equivariant Network}

\hypersetup{
pdftitle={Meta-Learning for One-Class Classification with Few Examples using Order-Equivariant Network},
pdfauthor={Ademola Oladosu, Tony Xu, Philip Ekfeldt, Brian A. Kelly, Miles Cranmer, Shirley Ho, Adrian M. Price-Whelan, Gabriella Contardo},
pdfkeywords={one-class classification, meta-learning, few shot learning, neural networks},
}

\begin{document}
\maketitle

\begin{abstract}
This paper presents a meta-learning framework for few-shots One-Class Classification (OCC) at test-time, a setting where labeled examples are only available for the positive class, and no supervision is given for the negative example. We consider that we have a set of `one-class classification' objective-tasks with only a small set of positive examples available for each task, and a set of training tasks with full supervision (i.e. highly imbalanced classification). We propose an approach using order-equivariant networks to learn a 'meta' binary-classifier. The model will take as input an example to classify from a given task, as well as the corresponding supervised set of positive examples for this OCC task. Thus, the output of the model will be 'conditioned' on the available positive example of a given task, allowing to predict on new tasks and new examples without labeled negative examples. 
\\
In this paper, we are motivated by an astronomy application. Our goal is to identify if stars belong to a specific stellar group (the 'one-class' for a given task), called \textit{stellar streams}, where each stellar stream is a different OCC-task. We show that our method transfers well on unseen (test) synthetic streams, and outperforms the baselines even though it is not retrained and accesses a much smaller part of the data per task to predict (only positive supervision).  We see however that it doesn't transfer as well on the real stream GD-1. This could come from intrinsic differences from the synthetic and real stream, highlighting the need for consistency in the 'nature' of the task for this method. However, light fine-tuning improve performances and outperform our baselines. Our experiments show encouraging results to further explore meta-learning methods for OCC tasks. 
\end{abstract}



\maketitle

\section{Introduction}

One-Class Classification (OCC), or class modelling, is a specific setup of machine-learning that tries to identify instances of a specific class among other objects, relying only on positive labeled examples, without negative supervision. This differs from the usual classification problem where one aims at distinguishing between different classes, with supervision for the different categories. This problem is also closely related to \textit{Positive and Unsupervised Learning} \cite{liu2003building,elkan2008learning} and \textit{transductive learning} \cite{joachims1999transductive}. The OCC setup is often considered in the context of anomaly or rare event detection: while there might be only a few anomalous examples (or if they are unknown), it might be easier to obtain `normal' examples, and model those as the positive class. In this setting, access to a reasonable number of `normal' examples is assumed. By modeling the normal class, one can then detect anomalous instances that diverge from that class.
\\
Some applications however face a similar problem at the extreme inverse, where only a handful of examples (a specific type of anomalies, or a rare-class) are known and labeled, and one wants to find additional instances of those in a pool of unlabeled examples. For instance, one could think of applications on co-expressed genes in genome, finding additional molecules with specific properties, information retrieval in a large dataset of documents, or as explored in this paper, finding stars from a specific stellar group. The goal here is thus to find `more needles in a haystack', with only a few needles examples, and where the haystack might contain a large number of irrelevant instances, and an unknown number of relevant ones, which in turn might make the creation of relevant and useful negative examples difficult. In this setting, the small amount of examples available to characterize the \textit{One-Class} will challenge existing methods for OCC framed for modeling the `normal' class.
\\
In this sense, this problem relates to few-shot learning, which focuses on the problem of learning with a small set of labeled examples, usually in the context of classification (binary or multi-class), e.g. with 3-5 examples per class. This problem has seen a renewed interest in the past few years through novel approaches using meta-learning. These methods propose to rely on a bigger `meta-dataset', composed of multiple few-shot-classification tasks. The idea is, since the original dataset for a single few-shot learning task is too small, to use a lot of these tasks to extract higher-level information and knowledge on how to solve the problem in a meta-fashion. Different types of meta-learning methods have been proposed, from optimization-based \cite{andrychowicz2016learning, nichol2018first, finn2017model,ravi2016optimization} to memory-based \cite{graves2014neural, weston2014memory}, as well as representation-based models \cite{snell2017prototypical, koch2015siamese, vinyals2016matching, sung2018learning}.
\\
Most few-shot learning methods focus on the multi-class case, often with fixed K-class. However, for many applications with actual few examples in `real life', such as rare-class classification or anomaly classification, framing the problem as One-Class Classification seems more sensible. Obtaining good negative examples (non-anomalous) might be difficult --as it is the goal of the task in the first place-- , and false-negative examples could dramatically impair the task at hand. This is motivated e.g. on medical data in \cite{irigoien2014towards}. Moreover, the `remaining examples' (unlabeled) will not fit the `few-shot' setting, as they will be likely in a high quantity. Finally, framing the problem as One-Class Classification, in a meta-learning setting, will allow for easier integration of new rare-class or types of anomaly at test-time. The motivations to use meta-learning for few-shot learning in order to leverage the information from a larger set of tasks are relevant for OCC as well: if one has access to a dataset of OCC tasks fully supervised, a meta-learning approach could extract how to efficiently model or characterize a class using only few examples. 
\\~\\
Therefore, we propose to address the problem of One-Class Classification with a meta-learning approach. We consider a meta-training dataset with fully supervised OCC problems (i.e. highly imbalance binary classification). The key idea of our model is to have an architecture that is trained as a binary classifier, but can be used at test-time on new tasks with only positive labeled examples and alleviate the need for negative supervision. To do so, we present a model that takes as input the set of positive examples in addition to the instance to classify. Namely, this model will (meta-)learn how to compute the distance metric and boundary from a set of positive examples and another instance, using a dataset of several such tasks of similar nature. The model therefore becomes a 'conditional' binary classifier, i.e. conditioned on the positive supervised set taken as input, and can predict on new tasks without seeing any negative examples. We propose to use order-equivariant networks to deal with the varying sizes of positive sets from one task to the next. 
\title{Meta-Learning for One-Class Classification with Few Examples using Order-Equivariant Network}
\\~\\
We present experiments on an astronomy application that motivated this setting and our approach. Our goal is to characterize the stellar populations of stellar streams (groups of co-moving stars that share some properties), where one stellar stream can be seen as an OCC task: we can determine a group of stars that we are confident belong to the stream (the positive class of the task), and we want to determine which other stars around belong or not to that stream. This problem is motivated by the importance of detecting sub-structures (e.g. gaps) in the shape of stellar streams, where therefore having a reliable population catalog is crucial. More information on the problem and the data are provided in Section \ref{sec:data}, and we provide the datasets created and used in this paper at \url{https://github.com/ReginaMayhem/ClusterBusters}.  Our experiments show that having such a meta-learned approach allows to alleviate the need for negative supervision: on test-tasks of similar nature (synthetic streams), our method outperforms baselines trained in a fully supervised regime on each class, i.e. baselines with a significant advantage as they access additional information (the negative supervision in an imbalanced binary classification regime) and are optimized for each single task (while our model is not fine-tuned but only used as a `forward'-pass). However, we observe a drop in the transferring ability of our approach when applied to a test-task of a slightly different nature (real stream). While fine-tuning (using both negative and positive supervision) helps to improve the results and outperform the baselines, such negative data might not be available in many cases. This high-lights the need for a meta-training set representative of the target (test) tasks, which may also not be straightforward for all applications. This work motivates further exploration of meta-learning methods in One-Class Classification regime, as they might prove crucial for many rare-class characterization problems in various fields. Exploration of more complex order-equivariant network architectures might also be of interest. 

\begin{figure*}
    \centering
    \includegraphics[width=0.75\linewidth]{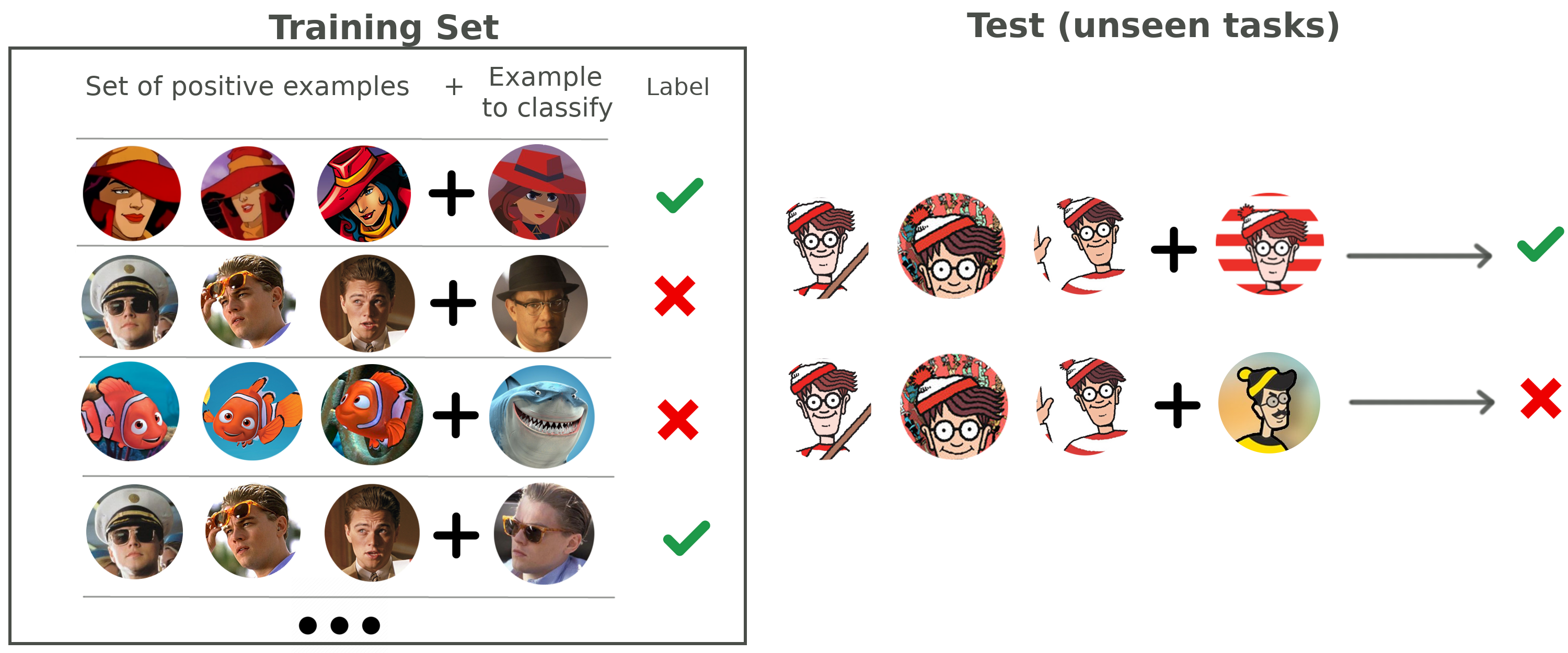}
    \caption{Illustration of our `meta'-training and testing setup: we train our model on a set of training instances, taken as input by our model. An instance is composed of (i) a set of positive example from a class $C$ and (ii) an example $x$ to classify, where the corresponding label is true if the example $x$ is from the class $C$ depicted by the positive examples, false otherwise. Those instances are generated from a set of $N$ tasks for which we have positive and negative examples. The model predicts on test instances built similarly, with a set of positive examples and an example to predict on, on classes unseen in the training set, i.e. it can predict on new tasks using only positive supervision. Note that, in practice, the sets of positive examples might be of different sizes for different instances and tasks. }
    \label{fig:waldo}
\end{figure*}

\section{Related Work}
\label{sec:rw}
~\\\textbf{Anomaly Detection and One-Class Classification}
~\\ 
Detecting or characterizing 'anomaly' has been a widely studied problem in machine learning and for various applications. We will present only briefly some references here, and refer the readers to surveys \cite{chandola2009anomaly} and  \cite{chalapathy2019deep}, the latter focusing on deep-learning methods. Anomaly detection is usually formulated as finding instances that are dissimilar to the rest of the data, also called outlier detection or out-of-distribution detection. It can be either unsupervised, supervised or semi-supervised.
\\
The `semi-supervised' case, or one-class classification \cite{moya1996network, khan2009survey} , considers that one only has one type of available labels (usually the 'normal' class as it is often easier to obtain in large quantities). The goal is usually to learn a model of the class behavior to get a discriminative boundary around normal instances. Counter-examples can also be injected to refine the boundary. Some of the techniques proposed are One-Class SVM \cite{scholkopf2000support,manevitz2001one, li2003improving} or using representation-learning with neural network \cite{perera2019learning}. This setting is also closely related to Positive and Unlabeled Learning (see e.g., \cite{liu2003building,elkan2008learning}).
\\
When supervised, the problem becomes largely similar to classic prediction with the main issues being getting accurate labels and the highly imbalanced data. The latter has been addressed e.g., through ensemble-learning \cite{joshi2002predicting}, two-phase rules (getting high recall first then high precision) \cite{joshi2001mining} or with cost-based / classes re-weighting in the classification loss. It is highlighted in \cite{chalapathy2019deep} that deep-learning methods don't fare well in such setting if the feature space is highly complex and non-linear.  
~\\~\\\textbf{Meta-Learning and Few-Shot Learning} 
~\\
Meta-learning aims at designing models able to learn how to learn, i.e., how to use information such as a small supervised dataset for a new, unseen task. The goal is to have a model that will adapt and generalize well to new tasks. To do so, the model will usually be trained on several similar tasks, with a 'meta'-training dataset. Various methods have been proposed with three main type of approaches:
(i) optimization based methods, which aim for instance at predicting how to update the weights more efficiently than usual optimization \cite{andrychowicz2016learning, nichol2018first, finn2017model,ravi2016optimization},
(ii) memory-based and model-based methods relying on models likes Neural Turing Machine \cite{graves2014neural} or Memory Networks \cite{weston2014memory} which are trained to learn how to encode, store and retrieve information on new tasks fast \cite{santoro2016meta, bartunov2019meta}, or are based on specific architectures with "slow weights" and "fast weights" dedicated for different parts of the training \cite{munkhdalai2017meta},
(iii) representation or metric learning-based methods, which aim at learning a good representation function to extract the most information from the examples from a task, in order to then use that representation as a basis to measure e.g., a linear distance with the unlabeled examples \cite{snell2017prototypical, koch2015siamese, vinyals2016matching, sung2018learning}. 
 \\We refer interested readers to more extensive surveys on meta-learning \cite{ vanschoren2018meta,hospedales2020meta} and few-shots learning \cite{wang2019few}.
 \\
 ~\\Meta-learning for OCC has only been very recently explored and, to the best of our knowledge, only a few methods have been presented. \cite{frikha2020few} proposed a modified Model-Agnostic Meta-Learning. A medical application is presented in \cite{gamper2020meta} using Deep One-Class method with meta-learning probabilistic inference, however they consider the `normal'-class (i.e. not rare) as the One-Class. Lastly, contemporary and closest to us,  \cite{dahia2020meta} proposed a representation-based approach combined with Support Vector Data Description (SVDD) \cite{tax2004support}, and an adaptation of Prototypical Networks \cite{snell2017prototypical} for OCC. Their meta-learning framework is closely related to ours, as they propose to adapt `many-classes' datasets from the few-shot literature to build binary one-class problems to meta-train on (note however in our application, or in many `rare-class' / anomaly applications, this might not be directly adaptable). They build their approach as a two-step strategy, first meta-learning a good representation space for the data for the OCC objective, and then solve the OCC tasks in this representation space. We note that \cite{sung2018learning} could be adapted within this framework as well. 
 \\The key difference with our approach is that we learn a representation transformation not per data-point, but of the entire set of examples describing a positive class, augmented by the example to classify (as being from this class or not). Therefore, we learn an order-invariant representation of the set \textit{conditioned} on the current example of interest (or vice-versa). Doing so allows us to bypass choices on the distance metric to use or choices to build a prototypical representation for the class (both being implicitly learned by the networks). 

\section{Model}
\label{sec:model}

We now present our proposed approach in more details. We first describe the learning setting, its objective and the training process in our context of meta-learning an OCC classifier, and then describe the proposed architecture.
~\\~\\\textbf{Learning}
~\\
%
Let us first define a \textit{training task} $T_j$ as composed of:
\begin{itemize}
    \item A \textit{support} set $P_j$ of $k_j$ positive examples 
    $P_j = \{ p_1^j, \dots, p_{k_j}^j\}$, with each example $p_i^j \in \mathbb{R}^n $
    \item A set $U_j$ of  $m_j$ supervised examples to classify $U_j = \{u_1^j, \dots, u_{m_j}^j\}$, with each example $u_i^j \in \mathbb{R}^n$ and their associated labels 
    $Y_j = \{y_1^j, \dots, y_{m_j}^j \}$, with $y_i^j \in \{0,1\}$
\end{itemize}
From a given task $T_j$, we can generate `paired instances' $x_i^j$, where $x_i^j$ is composed of an example to classify $u_i^j$ and its associated support set of positive examples $P_j$. This instance is labeled by $y_i^j$ (i.e. in a binary fashion). From an ensemble of training tasks (i.e. various $T_j$ with $j \in K$), we can generate a meta-training set of such binary-labeled paired instances, as illustrated in Figure \ref{fig:waldo}. Note that, in practice, one could sub-sample the support set for each task (i.e. use less than $k_j$ examples), thus creating many possible instances $x_i^j$ for a single example $u_i^j$, and/or generate more meta-training examples by resampling and shuffling the sets $P_j$ and $U_j$. 
\\~\\
We propose to denote a model $f$ with parameters $\theta$, which takes as input a paired instance $x_i^j = \{ u_i^j, P_j \}$, where $P_j$ can possibly be of varying size from one task $T_j$ to the next. We define this model $f_\theta$ as a binary classifier that outputs the Softmax probability for the instance $u_i^j$ to be from the positive class for the task, characterized by the support set $P_j$. In other words, the model $f$ can be seen as a binary classifier conditioned on the positive class directly through its input. Following this definition, $f_\theta$ can be trained by optimizing the cross-entropy loss empirically as:

\begin{equation}
       \min_{\theta} \sum_{j=1}^{K} \sum_{i=1}^{m_j} - \left[  y_i^j \log(f_\theta(P_j, u_i^j))  +  (1 - y_i^j)(1 - \log(f_\theta(P_j, u_i^j)))  \right]
\end{equation}
%
%
~\\~\\\textbf{Architecture}
\\In this framework, we propose to have a `representation-based' approach: at the difference of e.g. optimization-based methods, we do not aim to predict how to train fast, but instead we want to meta-learn a representation space and a metric measure that will transfer efficiently on any new tasks directly (i.e. without re-training).
\\Since we define an instance to classify on as the combination of the example to classify and the associated support set $P_j$ of positive supervised examples, we require an architecture that can handle such inputs of varying sizes (sets). We propose to rely on order-equivariant architectures: these methods originally designed for point-clouds can work on sets. Specifically, we use a Deep Sets architecture \cite{DS}. These architectures are two-folds: (i) one using order equivariant layers to build a fixed-size representation of the given set (or point-cloud), (ii) a secondary network (e.g. fully connected) to predict from this representation (in our case, the Softmax probability of the example to be in the same class as the support set).
\\
~\\
We propose to combine the support set of positive examples $P_j$ with the current example $u_i^j$ to classify w.r.t. to this support set, to be taken as input of the order-equivariant network directly. We re-factor the paired instance $x_{j,i}= \{P_j, u_i^j \}$ as the set of support examples concatenated to the example to classify: $x_i^j$ therefore becomes a set of $k_j$ examples in $\mathbb{R}^{2*n}$, as $x_i^j = \{ [p_1^j,u_i^j], \dots, [p_{k_j}^j, u_i^j]\}$, where $[p_k^j, u_i^j] \in \mathbb{R}^{2*n} $. This means that the primary network (composed of order-equivariant layers) builds a representation of a `point-cloud' in a feature space that is `conditioned' (i.e. already contains information) on the example to classify. While it would also be possible to have the primary network takes only the original set $P_j$ as input, and concatenate $u_i^j$ to the representation of $P_j$ in the secondary network (which would result in a similar approach as \cite{dahia2020meta} or a version of \cite{sung2018learning} adapted for OCC), we observed in practice that doing so was much less efficient. 
\\In our setting, this allows to learn a representation space conditioned on the support set and its relation to the instance to classify. It also allows to bypass any decision on the distance-metric to use (e.g. comparing a representation of the set \textit{versus} a representation of the instance) since this is handled by the secondary network. 
~\\~\\
\textbf{Prediction at test-time}
~\\By design, for a new task $T_j, j \notin \{1,K\}$, the model can classify any example $u_i^j$ needing only a set of positive examples $P_j$ for this task. It does so as a `forward-pass', without requiring additional training.


\section{Application and Data}
\label{sec:data}

This section provides information on our application of interest and details on the data used in this paper, as well as brief descriptions of physical meanings and motivations of the features. Formatted datasets used in our experiments are released at \url{https://github.com/ReginaMayhem/ClusterBusters}. 

\subsection{Description of the problem}
\begin{figure}
\centering
  \includegraphics[width=.2\linewidth]{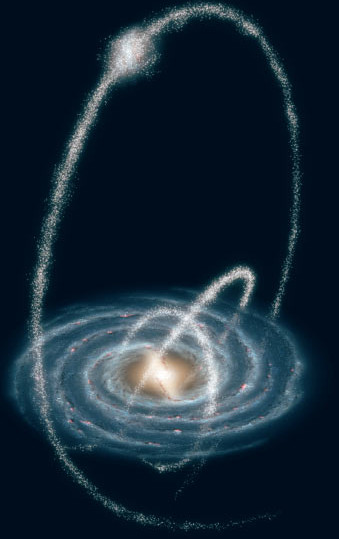}
  \caption{Artistic illustration of stellar streams wrapping around a galaxy.  Image credit: NASA / JPL-Caltech / R. Hurt, SSC \& Caltech. }
  \label{fig:illu_streams}

\end{figure}

\begin{figure}
\begin{minipage}{.47\textwidth}
  \centering
  \includegraphics[width=.9\linewidth]{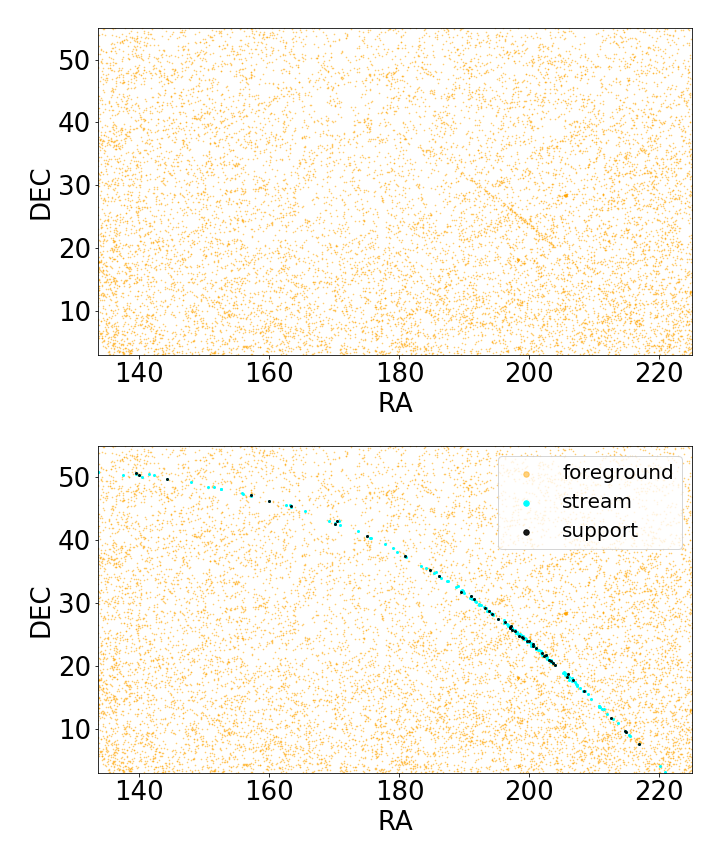}
  \caption{`Patch of the sky' around a synthetic stream, in 2-dimensions positions in the sky (RA-DEC), with foreground sub-sampled to a ratio 1:150. Bottom plot highlights the ground-truth with all the stream's stars in cyan, and a possible support set (a subset of the stream population) in black.} 
  \label{fig:radec_example}
\end{minipage}%
%
\hspace{0.06\textwidth}
%
\begin{minipage}{.47\textwidth}
  \centering
   \includegraphics[width=.9\linewidth]{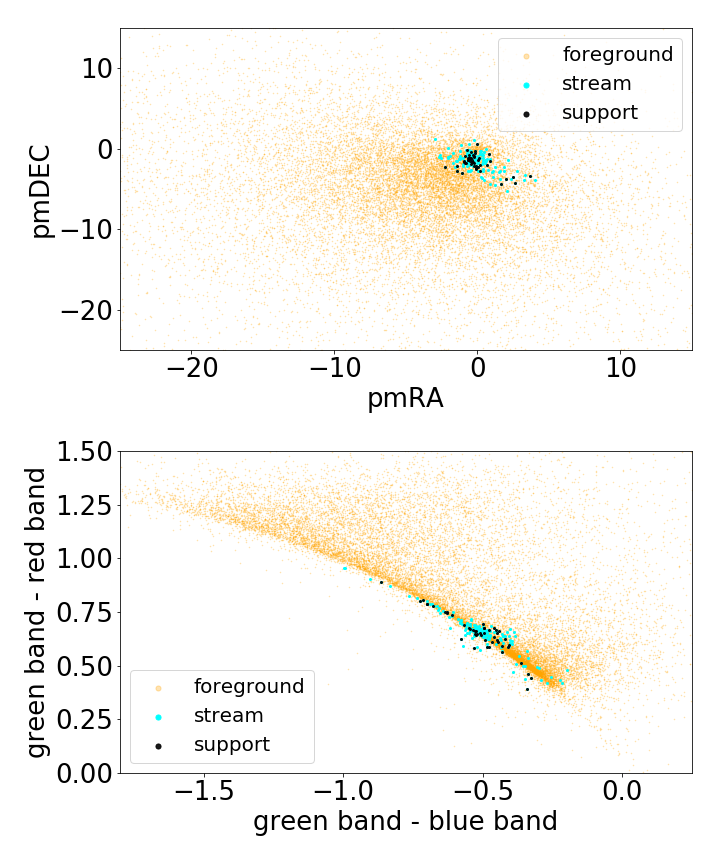}
  \caption{A synthetic stream (cyan), a possible support set (a subset of the stream population) (black) and foreground stars (ratio 1:150, yellow) in proper motion (pmRA-pmDEC) feature space (upper plot) and two of the colors features (bottom plot).}
  \label{fig:pmra_example}
\end{minipage}%

\end{figure}

We propose in this paper to focus on an astronomy application that aims at better characterizing the stars populations of \textit{stellar streams}. A stellar stream is a group of co-moving stars that orbits a galaxy. Stellar streams are thought to originate from smaller galaxies and star clusters that got accreted onto the main galaxy. These groups of stars are deformed and pulled apart by the differential gravitational field (the ``tidal field'') of the larger galaxy around which they orbit. The stars that are pulled out of these smaller systems end up forming linear, ``spaghetti-like'' structures of stars that continue to orbit coherently for (typically) several orbits around their parent galaxy. Figure \ref{fig:illu_streams} shows an artistic visualization of remnants of satellite galaxies wrapping around a bigger galaxy similar to the Milky Way, forming several stellar streams. Astronomers detect stellar streams in the Milky Way using large-area sky surveys that provide photometric (imaging), astrometric (sky motion), and radial velocity (line-of-sight motion) data. 
\\Typically, searches are performed by making cuts in specific regions of features space (looking at a specific region of the sky, looking at a specific type of orbit to select co-moving stars), and actual detection is usually done by visual inspection of 2D-visualizations of stars' positions in the sky, by searching for over-densities that are stream-shaped (see Figure \ref{fig:radec_example} which shows stars in 2D position in the sky dimensions, within a `patch of the sky', with a synthetic stream and the subsampled \textit{foreground} stars. Part of this stream forms a visible over-density in the top figure.). 
More recently, automated approaches have also been explored, e.g., \cite{StellarStreamDetection}, but even these methods typically require a significant amount of visual inspection and validation. 
\\
These streams are immensely useful tools for astronomers and astrophysicists: they are one of the most promising avenues for uncovering the astrophysical nature of dark matter and for inferring the accretion history of the Galaxy \cite{Kuepper:2015, GalaxyGhosts, AccretionHistory}. 
More specifically, sub structures within streams are of particular interest: for instance, gaps and spurs within a stream in the 2D sky's positions can be signatures of interactions with clumps of (otherwise invisible) dark matter \cite{banik2018probing, DarkMatterStellarStream}. In order to detect these structures however, one needs a reliable catalog of a stream's stellar population (i.e., determining which star belongs or not to a given stream). This will allow astronomers to analyze the internal substructure or  population characteristics of the stream. This is the problem we focus on in this paper.
\\
~\\
In this context, we consider a stellar stream as a task $T_j$, for which we can get a support set $P_j$ of positive examples, composed of stars belonging to the stream (for instance stars within the over-dense region for which we are confident). Our goal is to classify the remaining stars in the area, i.e. $U_j$, as belonging or not to this stream. Figure \ref{fig:radec_example} motivates this setup: it shows a `patch' of the sky, in 2-D space of positions projected on the sky, where each dot is a star (note that the data lies in more than these two dimensions, these features are described in the following Section and Figure \ref{fig:pmra_example} also illustrates this stream and its foreground in different feature space). On the top figure, one can distinguish by eye a stream-shaped overdensity (around 180-200 RA and 20-35 DEC). The bottom figure highlights which stars in this patch actually belong to this stream in cyan: one can see that the stream actually extends much further in the trajectory, however not in a perfectly smooth fashion. This also illustrates the potential difficulty to pick informative negative examples. This motivates a \textit{One-Class} approach. Note that the \textit{foreground} (i.e. the stars that are not from this stream) is already sub-sampled to mimic the actual data one might observe after detecting a stream (which would otherwise not be visible by eye). Following the notations introduced in the previous section, a training task $T_j$ would have access to all the labels from the bottom figure, i.e. 0 for the foreground stars and 1 for the stream stars, in addition to the support set of positive stream stars $P_j$, shown in black. At test-time however, i.e. if we observe a new stream, the only supervision available are the stars from the support positive set for the new stream i.e. the black points in Figures \ref{fig:radec_example} and \ref{fig:pmra_example}. In practice in this paper, the support sets for the synthetic streams used in training, validation and test, are sampled randomly across each stream's population, and not restricted to a specifically over-dense region in RA-DEC (we provide more details on the dataset building in Section \ref{sec:exp}): as shown in Figure \ref{fig:radec_example} some of the black points extend beyond the over-dense stream-shaped region in RA-DEC. 


\subsection{Data}
As explained above, our targeted problem is to characterize the stellar populations of detected streams, where for each stream we consider we have access to a reliable \textit{support} set of positive stars belonging to that stream (i.e. a sub-set of the entire true population of stars of this stream) as sole supervision to classify the rest of the stars. For almost all currently known real streams, there is no extensive or complete catalog available for these populations. However, it is possible to simulate synthetic streams: those synthetic streams could allow us to have full ground-truth and thus supervision, in order to extract higher-level information on the `meta distance metric' that characterize the membership of a star to a given stream (based on a sub-set of its stars). We propose to use such synthetic streams in this paper to meta-train and evaluate our approach with full ground-truth. While those streams are synthetic, we inject them in real data (i.e. non synthetic `foreground' stars), which is the same real data that will be used down the line to characterize the population of real streams, from the Gaia mission. We also provide preliminary results on one real stream, GD-1, for which we have a reliable catalog. 
The following provides more details on the stars' features in our datasets, as well as the datasets for the `real' stars (foreground and GD-1) and for synthetic streams.

~\\\textbf{Features}: Each star in our datasets have the following ten features: 

\begin{itemize}
        \item \textbf{RA-DEC Position in the sky}: Positions of the stars projected in the Equatorial Coordinate System, in two dimensions: \textit{RA} is the barycentric right ascension; \textit{DEC} is the barycentric declination. The characteristic shape of the stream will be observable in this 2D space (e.g. Figure \ref{fig:radec_example}). 
        
        \item \textbf{Proper motions}: Movement of an astronomical object relative to the sun frame of reference, \textit{pmRA} in the direction of right ascension; \textit{pmDEC} in the direction of declination. The stream's stars will also be structured in this 2D space as they share common motion properties from their orbit around the Milky Way, as illustrated in upper plot in Figure \ref{fig:pmra_example}.
        
        \item \textbf{Colors}: Each star in the data set has three photometric features: the mean absolute magnitudes for the green band, the green band minus the red band, the green band minus the blue band. These features are indirect indicators of the potential age and composition of a star\footnote{There exists a (non-linear) relationship between stars' ages and observed colors called an isochrone.} and we expect stars from a stream to have roughly similar ages. Bottom plot in Figure \ref{fig:pmra_example} shows a stream and its foreground in 2 of this 3-D color space. 
        \item \textbf{Angular coordinates}: We additionally use the angular velocity coordinates of each star, which combines proper motions and the equatorial coordinates. Essentially, these angles represent the great circle along which the star is moving across the sky.
    \end{itemize}
%
%
~\\\textbf{\textit{Gaia} data}:  
\textit{Gaia} is a space observatory designed for astrometry. This mission has produced a dataset of stars in our Milky Way of unprecedented caliber. The spacecraft measures the positions, distances and motions of stars in the Milky Way brighter than magnitude 20, which represents approximately 1\% of the Milky Way population. The mission also conducts spectrophotometric measurements, providing detailed physical properties of the stars observed, such as luminosity, and elemental composition. Recently, the second dataset of observations (\textit{Gaia} DR-2  \cite{GaiaDocumentation}) has been released. It contains measurements of the five-parameters astrometric solution -- positions on the sky, parallaxes, and proper motions in two dimensions -- and photometry (colors) in  three bands for more than 1.3 billion sources. All of our \textit{foreground} stars (i.e. `negative' examples) for each training task $T_j$ are from Gaia data, as well as GD-1's stars.
\\
~\\\textbf{Synthetic streams}: 
The synthetic streams are generated by simulating a collection of star clusters as they orbit the Milky Way. 
Star `particles' are ejected from a mock (massive) star cluster, and the orbits of the individual star particles are then computed accounting for the mass distribution of the Galaxy along with the mass of the parent star cluster.
The star cluster orbits are randomly sampled to match a plausible initial radial distribution of star clusters born or accreted into the Milky Way.
These simulations are performed with the \texttt{Gala} Galactic Dynamics code \cite{Price-Whelan:gala}. \\
After evolving the orbits of the star particles ejected from all of the individual star clusters, the final state of these simulations is a set of synthetic stellar streams: Positions and velocities in a Galactocentric reference frame for all star particles in all synthetic streams. 
We then transform these positions and velocities to heliocentric quantities and mock observe the star particles to mimic the selection function and noise properties of sources in \textit{Gaia} DR-2.
These streams are then superimposed over the real \textit{Gaia} data: because the streams are generated so as to orbit around our actual galaxy, we can mimic the "foreground" we would observe if those stream were real (i.e., in terms of positions in the sky / equatorial coordinates). 
Thus, we can generate realistic datasets composed of real data from \textit{Gaia} (for the foreground, `negative' class for each stream) and synthetic streams, where we have supervision (ground truth) for all stars. 
Each stream dataset is generated by selecting a random window in RA-DEC that contains part of the stream, and injecting real foreground stars from \textit{Gaia} to a ratio of up to 1:150 \footnote{This has been estimated based on the ratio in known streams after astronomically relevant cuts are performed.} with galactic disk removed. 

~\\\textbf{GD-1}:
We also show results for one real stream, GD-1, for which we have an exhaustive catalog based on astronomical cuts described in \cite{GD1, DarkMatterStellarStream}.
~\\~\\
The following section provides details on our experimental protocol and how we use these data to generate our meta-training and testing sets.

\section{Experiments}
\label{sec:exp}

Dataset and code (for baselines and Meta-Deep Sets as well as trained networks to reproduce results) are available at  \url{https://github.com/ReginaMayhem/ClusterBusters}.

\subsection{Baselines} 
Our target problem and application is in the One-Class Classification setup at test-time: we aim at classifying a set of unlabeled examples for a task where we only have supervision for the positive class. However, preliminary experiments using one-class methods such as One-Class SVM showed that these approaches performed very poorly. This might be explained by the relatively small sizes of the positive training sets for the tasks (between $\sim$ 10 and $\sim$ 200 examples) and the high imbalance regime. 
\\Therefore, we propose to compare our method to baselines trained in the classic binary classification setting for each test task. This means that, for a given test-task, the baseline has access to much more information than our method, because it uses \textit{negative} supervision, which our model doesn't see. Additionally, for each test task $T_j$, the baselines are optimized on this single task, while our approach is not re-fitted or fine-tuned on new tasks, but only predicts as a forward pass (except when explicitly mentioned, see experiments on GD-1). Thus, the baselines have a significant advantage `locally' for each task. However, they do not access higher-level information from the larger set of tasks (except for hyper-parameters choices). These experiments will highlight if there is higher level, `meta' information, that can be extracted from a set of training tasks of similar nature and if this can alleviate the need for negative examples on new classes (new tasks) at test time, and thus showcase the interest and advantage of a meta-trained model. 
\\~\\
Specifically, we propose to use Random Forests as a baseline, as they provide usually good performances over a variety of machine learning problems. They are also robust to dataset with imbalanced supervision. 
 Our preliminary experiments showed that the Random Forest (RF) approach could achieve relatively high precision with medium or low recall (i.e., conservative models with few false positives) for some tasks. Given that one of the challenges of the problem at hand is the low number of positive examples compared to the negative ones, we propose to also explore self-labeling \cite{triguero2015self} with RF. The idea is to use predicted labels from the model to augment the set of (training) positive examples, and retrain. One can start with an initial Random Forest that is highly conservative (high precision / low recall), which means 'safe' examples but probably less informative are added to the training set, or a more balanced mixture between precision and recall.  Our validation-protocol indicates better results when using such criteria (e.g., F1) with medium precision and recall, which are the results shown in this paper. We can repeat the self-labeling process for a certain number of steps or until a stopping criterion (e.g., no positive examples are predicted anymore in the self-labeling pool). The following subsection also describes the split used to keep a pool of separate examples for self-labeling.
\\
Hyper-parameters such as the number of iterations, but also number of trees in forest (100, 200, 300, 500), max depth (10,30,50), min split (2,5,10), min leaf (1,2,4), bootstrap vs. not bootstrap, are selected in a meta-validation fashion (i.e., on a group of streams tasks used for validation,  described below), i.e. all RF models fitted on the test-tasks share hyper-parameters. 

\subsection{Data Pre-Processing and Dataset building}
\label{section:data:split}
We build a meta-dataset of 61 synthetic stellar streams and their respective foreground. We split them into a meta-training, meta-validation and meta-test dataset composed of respectively 46, 7 and 8 streams tasks. Each "stream dataset" has a ratio of 150 negative examples for 1 positive example. 
~\\~\\\textbf{Meta-training dataset} From each train-stream dataset, we generate training instances composed of (i) a support set of varying size randomly sampled within the stream's stars, (ii) a star to classify, (iii) its corresponding label. We can generate meta-training datasets with varying imbalance between negative and positive instances, by changing the ratio of negative examples used and by duplicating positive instances by sampling different support sets (i.e. the example to classify is the same, but the support set is sub-sampled differently). Results shown here are from models trained on a meta-training dataset with a final imbalance of 1:50 positive vs negative instances. This dataset is not used by the baselines.
~\\~\\
\textbf{Meta-Validation and Test datasets}
For each task in the validation and test tasks, we use 10\% of the stream's stars as "training examples" --used to train the RF, or used only as the support set $P_j$ taken as input by our approach--. In the Meta-Test set, the smallest support set (resp. biggest) has 9 stars (resp. 92 stars). Average size of the support set is 36 examples. The RF also has access to 10\% of the foreground stars (negative examples) to train on (i.e keeping a ratio of 1:150). Our approach does not have access to those stars by default. The remaining stars of each dataset are split into two groups, one for the self-labeling pool for the RF, the other half (\textit{final test}) for all final evaluation, common to all methods, to make comparison of results consistent. GD-1 dataset is built similarly with a support set (positive training examples, 197 stars), a negative training set (ratio 1:400), a 'self-labeling' set, and a test set (ratios 1:150 for both).
\\
Each task' dataset within Meta-Train, Meta-Validation and Meta-Test are normalized 'locally,' i.e., per task/stream, using all examples within the dataset of the task.
\\
Our approach is trained on the meta-training dataset. Validation and model selection is conducted on the meta-validation set. Our approach is not fine-tuned or retrained on the meta-validation or meta-test set. We also use the meta-validation dataset to select hyper-parameters for the Random Forest models. 
The following results are obtained for Deep Sets with 5 layers of size 100, and exploring the following hyper-parameters: learning rate $\{10^{-3}, 10^{-4}\}$, $l1$-regularization $\{10^{-5}, 10^{-6} \}$ and weight imbalance for the loss $\{0.01, 0.1, 1, 10, 100\}$.
\begin{table*}[t]
  \caption{Results on Meta-Test synthetic streams and GD-1 for Random Forest (RF), Random Forest with Self-Labeling process (RF Self-Lab), our meta-learning approach based on Deep Sets (DS), Deep Sets Fine-Tuned (DS FT). \textit{Best meta F1} indicates the selection criteria used on the Meta-Validation set. \textit{Best train F1} indicates criteria selection used on GD-1 training-set when fine-tuning. }
  \label{tab:sum_all}
\centering
  \begin{tabular}{l l  l l l l l l l} 
    Dataset & Model & Precision & Recall & F1 & F2 & F0.5 & BAcc & MCC \\
    \hline \hline
        \multirow{4}{*}{Synthetic} &
     RF & 0.590 & 0.538 & 0.499 & 0.502 & 0.534 & 0.766 & 0.526 \\  \cline{2-9} 
     & RF Self-Lab &  0.525 & 0.690 & 0.553 & 0.609 & 0.529 & 0.841 & 0.574 \\   \cline{2-9} 
     & Meta DS (best meta F1) &  \textbf{0.698} & 0.643 & \textbf{0.652} & 0.643 & \textbf{0.674}  & 0.821 &  \textbf{0.660}\\ \cline{2-9} 
     & Meta DS (best meta F2) & 0.502 & \textbf{0.859} & 0.619 & \textbf{0.737} & 0.541 & \textbf{0.927} & 0.646   \\  \hline \hline
     
     \multirow{8}{*}{GD-1} &
     RF & 0.739 & 0.575 & 0.647 & 0.601 & 0.699 & 0.787 & 0.650 \\  \cline{2-9} 
     & RF Self-Lab & 0.402 & 0.869 & 0.550 & 0.705 & 0.450 & 0.930 & 0.587 \\   \cline{2-9} 
     & Meta DS (best meta F1) & 0.490 &  0.473 &  0.481 &  0.476 &  0.486 &  0.735   &  0.478 \\ \cline{2-9} 
     & Meta DS (best meta F2) & 0.266 &  0.561 &  0.360 &  0.459 &  0.297  &  0.775 &   0.380 \\  \cline{2-9} 
    & DS FT (best train F1) &   0.731 &  0.675 &  0.702 &  0.686 &  \textbf{0.719} &  0.837 &  0.701 \\  \cline{2-9} 
    & DS FT (best train F2) &   0.624 &  0.834 &  \textbf{0.714} &  \textbf{0.782} &  0.658 &  0.916 &  \textbf{0.720} \\  \cline{2-9} 
    & DS FT (best train F0.5) &    \textbf{0.774} &  0.541 &  0.637 &  0.575 &  0.712 &  0.770 &  0.645 \\  \cline{2-9} 
    & DS FT (best train Rec) &   0.341 &  \textbf{0.981} &  0.506 &  0.713 &  0.392 &  \textbf{0.984} &  0.574 
     \\  
  \end{tabular}
\end{table*}
~\\~\\\textbf{Performance criteria}
Given the imbalanced nature of our classification task, we study the performance of the methods on several criteria that give various trade-off between precision and recall. The different measures are Precision, Recall, F-1, F-2 (favors recall), F-0.5 (favors precision), Balanced Accuracy (BAcc) and the Matthew's Correlation Coefficient (MCC) --or the Phi Coefficient--. BAcc is computed as $(\textrm{True Positive Rate} + \textrm{True Negative Rate} ) / 2$, where the True Positive Rate is the recall, and the True Negative Rate is the specificity (or the number of True Negative divided by the total number of negatives). BAcc 
assumes the cost of a false negative is the same as the cost of a false positive. MCC is also a balanced measure even if the class are of different sizes; it combines the four elements of the confusion matrix \cite{powers2011evaluation}. All scores shown here are computed by averaging the scores obtained for each stream task within the meta-test set, on the \textit{final test} stars.

\subsection{Results}
 ~\\\textbf{Synthetic Streams}
Upper part of the Table \ref{tab:sum_all} summarizes the score of our models selected to maximize different criteria. The hyper-parameters selected for the RF and RF with self-labeling were the same for F-1, F-2, MCC and Balanced Accuracy. The hyper-parameters selected for our approach (\textit{Meta DS}) led to two different models that maximized either $\{$F-1, F-0.5, MCC$\}$ (and precision) or $\{$F-2, BAcc$\}$ (and recall). Therefore we show both models on the Table. We see that our model manages to generalize well to new tasks, as it gets the best results for all criteria considered, with significant gain for all of them. This illustrates the ability of our approach to transfer efficiently to new, unseen tasks, without retraining or fine-tuning, and using only small, partial `supervision', on only the positive class. This high-lights that using meta-training on different tasks of similar nature alleviates the need for negative examples for new tasks, and even provides better results than directly fitting on the given task with more supervision. However, we note that there are still room for improvement to alleviate the (expected) trade-off between precision and recall, though we remind the reader that each task requires to accurately find on average about a hundred stars in a pool of 20 thousands examples.
~\\
~\\
\textbf{GD-1}
The lower part of Table \ref{tab:sum_all} shows the scores for GD-1 data for the models selected through the same validation process as the synthetic streams. We see that, on this real stream dataset, our Meta Deep Sets model struggles to obtain as good results. Comparatively, the RF and RF with self-labeling perform very well (but again, these baselines access more supervised data --negative examples-- and are optimized for this task). We see different factors that could explain the difference of results between synthetic streams and GD-1 with our meta-approach. First, the synthetic streams may be different in nature to GD-1. The Meta-approach seems to generalize well on new synthetic streams, but not here: this could highlight that either the synthetic streams are not realistic enough, or GD-1 has something inherently different. In particular, we know that GD-1 has a very distinctive orbit (observed in pm-RA/pm-DEC space); maybe those features are less important on the synthetic streams, or used differently in the 'meta-distance metric'. Since the RF is trained only on GD-1, it can pick this more easily. We also noted a relatively high variability in RF's performances on the synthetic test-streams (often in opposite direction to performance of our approach): maybe GD-1 is a type of streams that is `easier' to characterize (e.g. through its distinctive orbit) compared to others. 
\\We propose to briefly explore the use of fine-tuning on our approach, on the Meta Deep Sets with best meta-validation F-1 score. We use the same negative training set as the RF, combined with the original positive support set (also used by RF). We divide the learning rate by two. We modify the learning scheme so that at each fine-tuning epoch, the model is trained on a dataset sampled from the support set and $N$ negative stars, where $N$ is the number of positive examples $\times$ an imbalance factor. We try 4 imbalance factor (30, 50, 70 and 100). All fine-tuned models are retrained so as to see the entire negative dataset 3 times. We show results on the final test data (last rows of Table \ref{tab:sum_all}, where DS FT is Deep Sets Fine-Tuned) when selecting the best models for criteria f1, f2, f05 and recall on the GD-1 training-set. Fine-tuning leads to great improvement and allows to improve against baselines (showcasing the gain of extracting meta-information from other OCC tasks even if they are slightly different --here synthetic--). It also shows the potential of the model to reach a variety of trade-offs between precision and recall. However, these fine-tuned results are more shown as a proof of concept since, in practice, access to relevant negative examples might be difficult or impossible. 


\section{Discussion}
We presented in this paper a meta-learning approach for few-shots One-Class Classification tasks at test-time, i.e. tasks where the only supervision available is a small set of examples from the positive class, and no supervision for the negative class. We proposed a model using order-equivariant networks with a representation-based meta-learning approach. Our experiments on an astronomy application show the advantage of using a meta-learning setting for this problem, as we observe a significant gain in performance compared to baselines, even-though the baselines have access to more data (supervision on both classes) and are fitted for each test-task, compared to our model which only sees positive examples and is not re-trained or fine-tuned. However, we observed that our model struggles to perform as well on a `real stream' test-task: this could be explained by differences in properties between the synthetic streams and the real GD-1 stream. This high-light the need for a meta-training dataset that is representative to the target task at hand. These experiments however provide encouraging results that motivate further exploration of meta-learning methods for the One-Class Classification regime, as they could prove useful for many applications where the goal is to find more instances of rare classes of interest. Using more complex order-equivariant architectures could also be of interest for representation-based approaches to meta-learning.

\bibliographystyle{apalike}
\bibliography{example_paper}

\end{document}